\documentclass[sn-chicago]{sn-jnl}

\usepackage{graphicx}%
\usepackage{multirow}%
\usepackage{amsmath,amssymb,amsfonts}%
\usepackage{amsthm}%
\usepackage{mathrsfs}%
\usepackage[title]{appendix}%
\usepackage{xcolor}%
\usepackage{textcomp}%
\usepackage{manyfoot}%
\usepackage{booktabs}%
\usepackage{algorithm}%
\usepackage{algorithmicx}%
\usepackage{algpseudocode}%
\usepackage{listings}%
\usepackage{etoolbox}

\usepackage{multirow,booktabs,setspace,caption}
\usepackage{tikz}

\DeclareCaptionLabelSeparator*{spaced}{\\[2ex]}
\theoremstyle{thmstyleone}%
\usepackage{longtable}

\theoremstyle{thmstyletwo}%
\usepackage{booktabs}
\usepackage{appendix}

\theoremstyle{thmstylethree}%

\begin{document}

\title[A Research Study on LLM application for Automatic Assessment]{Fine-tuning ChatGPT for Automatic Scoring}


\author[1]{\fnm{Ehsan} \sur{Latif}}\email{ehsan.latif@uga.edu}

\author*[1]{\fnm{Xiaoming} \sur{Zhai}}\email{xiaoming.zhai@uga.edu}

\affil*[1]{\orgdiv{AI4STEM Education Center}, \orgname{University of Georgia}, \orgaddress{\city{Athens}, \state{GA}, \country{USA}
}}


\pretocmd{\abstractname}{}{}{}
\abstract{
This study highlights the potential of fine-tuned ChatGPT (GPT-3.5) for automatically scoring student written constructed responses using example assessment tasks in science education. GPT-3.5 has been trained over enormous online language materials such as journals and Wikipedia; however, direct usage of pre-trained GPT-3.5 is insufficient for automatic scoring as students do not utilize the same language as journals or Wikipedia, and contextual information is required for accurate scoring. All of these imply that a fine-tuning of a domain-specific model using data for specific tasks can enhance model performance. In this study, we fine-tuned GPT-3.5 on six assessment tasks with a diverse dataset of middle-school and high-school student responses and expert scoring. The six tasks comprise two multi-label 
and four multi-class 
assessment tasks. We compare the performance of fine-tuned GPT-3.5 with the fine-tuned state-of-the-art Google's generated language model, BERT. The results show that in-domain training corpora constructed from science questions and responses for BERT achieved average accuracy = 0.838, SD = 0.069. GPT-3.5 shows a remarkable average increase  (9.1\%) in automatic scoring accuracy (mean = 9.15, SD = 0.042) for the six tasks, \textit{p} =0.001 $<$ 0.05. Specifically, for each of the two multi-label tasks (item 1 with 5 labels; item 2 with 10 labels), GPT-3.5 achieved significantly higher scoring accuracy than BERT across all the labels, with the second item achieving a 7.1\% increase. The average scoring increase for the four multi-class items for GPT-3.5 was 10.6\% compared to BERT. Our study confirmed the effectiveness of fine-tuned GPT-3.5 for automatic scoring of student responses on domain-specific data in education with high accuracy. We have released fine-tuned models for public use and community engagement.
}







\maketitle
\newpage
\section{Introduction}
\label{introduction}
The importance of education in stimulating creativity, encouraging critical thinking, and establishing the groundwork for individual and societal improvement has long been acknowledged. This claim becomes more relevant today, marked by extraordinary technological advancements, a developing digital ecosystem, and a complex globalized society \citep{linn2014computer}. There has never been a greater need for a strong and effective education system that provides students with the knowledge and skills they need to successfully traverse the challenges of the 21st century. Despite the needs, modern educational frameworks sometimes find themselves up against complex problems. One of these is adaptability, as conventional teaching and learning methodologies occasionally find it challenging to keep up with the rapidly changing state of knowledge and technology. The challenge of personalization is particularly challenging--  a one-size-fits-all strategy \citep{teasley2017student} frequently fails to provide meaningful and unique learning experiences given the variety of student backgrounds and demands \citep{limna2022review}. 

Artificial intelligence (AI) has taken center stage in this dynamic environment as a possible game-changer for the education industry \citep{zhai2023gamechanger,latif2023artificial,lee2023multimodality}. AI promises to address some of the most challenging issues in education thanks to its unmatched data processing powers, predictive analytics, and adaptable algorithms. AI has the potential not just to augment but also completely transform current educational practices, whether it be through the creation of personalized learning pathways, the provision of real-time feedback, or even the facilitation of massive online courses with thousands of students \citep{namatherdhala2022comprehensive, limna2022review, zhai2020substitution}. The nexus of AI and education creates opportunities for more individualized, adaptable, and inclusive learning experiences, ushering in a new era of pedagogy \citep{holmes2022state, wu2023matching, zhai2023technology}.   Although the use of AI in education is growing, not all AI solutions are suited to the particular needs of the educational field. Generic AI models frequently lack the precise contextual awareness required for successful educational interactions \citep{zhai2021review,liu2023context, tahiru2021ai, ng2023review}.

The ChatGPT program, created by OpenAI, is a leading example of natural language processing and the promise of AI in education \citep{mhlanga2023open,zhai2023chatgpt}. While ChatGPT demonstrates outstanding broad conversational skills, there are inherent limitations when using it without adjustments in specialized domains like education \citep{zhai2022chatgpt}. The general trend towards fine-tuning AI technology for certain activities resonates with improving ChatGPT for educational purposes \citep{liu2023context}. A fine-tuned and applying Chain-of-Thought \citep{lee2023applying} using ChatGPT may offer individualized, engaging, and successful learning experiences consistent with the philosophy of lifelong learning by integrating domain-specific knowledge, context awareness, and pedagogical techniques \citep{zhai2022chatgpt}.

Inspired by the capabilities of ChatGPT in predicting sentences even with zero-shot \citep{wei2023zero} and remarkable responses with few-shots \citep{dai2023chataug,liu2023chatgpt} for different applications, we propose a fine-tuned ChatGPT for science educational automatic scoring. In this study, we fine-tuned the public version of ChatGPT, i.e., GPT-3.5, with six complex assessment tasks in science education. Our findings show high accuracy in automatic scoring using fine-tuned ChatGPT compared to the performance of Google's pre-trained language model BERT \citep{devlin2018bert}. This study can set grounds for future uses of fine-tuned Chatbots in education and shows high potential for science education assessments.

The study contributes to the field in multiple aspects:
\begin{itemize}
    \item We propose a GPT-3.5-tubo's fine-tuning scheme that utilizes a simple and explainable principle to select and obtain high accuracy for automatic scoring. This approach strives for both effectiveness and portability in evaluating and curating data subsets.
    \item We collected empirical evidence for how accurate the fine-tuned GPT-3.5-turbo is in scoring students' written responses to science assessments. Specifically, the study provides evidence for both multi-label and multi-class assessment tasks.
    \item This study compared the fine-tuned GPT-3.5-turbo model with a state-of-the-art language model (BERT) and showed the outperformance of GPT-3.5-turbo backed by the accuracy results.
    \item This study shows the potential of fine-tuned GPT-3.5-turbo in addressing unbalanced training data.
\end{itemize}

\section{Automatic Scoring in Education}
\label{automatic_scoring}

In the educational field, automatic scoring has become a cutting-edge approach to evaluating student-generated content without manual grading. This strategy is especially useful in large-scale classrooms when manual scoring is impractical \citep{susanti2023automatic, hahn2021systematic}. A critical development in learning technologies leverages the powerful machine learning algorithms \citep{teasley2017student, zhai2020applying}. These methods were designed to provide timely information for teachers to make informed instructional decisions, as well as provide customized feedback to students to guide self-regulated learning \citep{gerard2016using}. Student-centered learning theories also started to be incorporated into the design of automatic scoring \citep{matcha2019systematic}.

An earlier attempt at automatic scoring concentrated on short responses. For example, the c-rater was introduced in \citep{sukkarieh2009c} to score the content of such responses. The algorithm attempted to assess the quality and relevance of the content using sophisticated linguistic techniques. Researchers started looking into more complex scoring algorithms as the number and complexity of student comments increased, particularly in the context of classroom assessment practices. Aspect-level sentiment analysis was used by \citep{ren2023automatic} to grade student input automatically. Their strategy allowed for a more detailed evaluation of the feedback by classifying feelings following particular facets of training. There have also been developments in the area of automatic question-generating. The rising research in this field, which has the potential to supplement automatic scoring by creating standardized, AI-generated questions, was noted in a thorough assessment by \citep{kurdi2020systematic}.

Additionally, recognizing the inherent variety in student responses, \citep{horbach2019influence} looked into how this variability affected the automatic content scoring. Their study emphasized the significance of considering the many methods by which students express their ideas. In the context of essay exams, a systematic evaluation by \citep{susanti2023automatic} demonstrated a variety of approaches, including the use of neural networks and natural language processing, to grade lengthy responses. Meanwhile, \citep{fu2020affordances} studied the advantages of AI-enabled scoring applications in promoting students' intent to study continuously and found good correlations.

Although these developments are encouraging, there are still challenges to address. For example, \citep{zhai2021meta} conducted a meta-analysis of automatic scoring and reported varied success in scoring accuracy and identified a number of factors that contribute to machine scoring accuracy. The expectations of the automatic scoring system and the actual material are frequently out of sync due to the huge diversity of student-generated content \citep{horbach2019influence}. Additionally, it's still difficult to ensure fairness and eliminate biases, particularly in aspect-level sentiment studies \citep{ren2023automatic}. Another issue that limits the general applicability of scoring systems is their flexibility across different disciplines and languages \citep{susanti2023automatic}. 

The potential of fine-tuning language models emerges as a possible option in the face of these difficulties. Automatic scoring can be made more precise, adaptive, and context-aware by utilizing the extensive knowledge base and contextual understanding of models like ChatGPT and adding domain-specific expertise \citep{kasneci2023chatgpt}. This has the ability to overcome the current drawbacks and pave the way for an enhanced and all-encompassing automatic evaluation system for use in education. This study captures the high volume of student written responses and processed data for better results from fine-tuning ChatGPT; further, we also gather data from different demographics such as gender and race and compile the data in such a way to overcome the possible biases in automatic scoring.

\section{Large Language Models for Automatic Scoring}
\label{llm_for_scoring}
Large language models (LLMs) have drawn increasing attention in automatic scoring because of their extensive knowledge base, context awareness, and flexibility. Different from traditional AI models, LLMs are pre-trained using large datasets. For example, Google pre-trained a Transformer model BERT using Wikipedia data, which includes 12 encoders with 12 bidirectional self-attention heads totaling 110 million parameters in its base variant and can conduct various tasks such as word sense disambiguation, natural language inference, and sentiment classification.

Automatic scoring of student-written responses is a common application of LLMs. It was shown in \citep{organisciak2023beyond} that LLMs can efficiently evaluate divergent thinking problems by considering factors other than semantic distance. \citep{bertolini2023automatic} used LLMs to assess the emotional content of dream reports, demonstrating the variety of applications in a more specialized application. LLMs have proven to be reliable reviewers for traditional essay scoring, a field where machine evaluation has historically been difficult owing to the range and complexity of human expression. Transformer-based models are used for this purpose in works by \citep{rodriguez2019language} and \citep{ormerod2021automated}, demonstrating LLMs' effectiveness in grading lengthy text-based responses.

Even in specialized disciplines like mathematics and science education, there is an increasing trend towards using LLMs. For instance, \citep{liu2023context} and \citep{shen2021mathbert} explore BERT tactics specific to assess math and science learning, respectively. Additionally, \citep{wu2022automatic} suggested adopting LLMs for automatic grading in the translation domain, highlighting the models' ability to understand linguistic nuance.

Despite their abilities, LLMs face challenges in educational settings, such as concerns about prejudice and justice \citep{yan2023practical,zhai2022pseudo}. These models may unintentionally reinforce preexisting biases because they are trained on large datasets. Furthermore, due to their generic character, they frequently miss out on the nuances of a specific topic \citep{caines2023application}. The ability to interpret the predictions or classifications made by these models is another possible challenge. The 'black-box' aspect of some LLMs can be a source of dispute because stakeholders in education frequently require transparency and openness in grading \citep{kasneci2023chatgpt}.

Moreover, most of the LLM-based automatic scoring models are based on BERT \citep{devlin2018bert}, which has limitations despite being groundbreaking in its bidirectional approach to contextual embeddings. BERT is primarily intended for tasks that call for a fixed-length input, making generating sequences or handling more open-ended tasks more difficult. Because of its architecture, it is necessary to utilize laborious tokens like \textcolor{black}{[SEP] (a separator token) and [CLS] (a classification token), which will be used to predict whether or not second part (A) is a sentence that directly follows first part (B).} Additionally, BERT is not naturally generative, even though it can encode context in both directions. However, student-written responses are not fixed-length and sometimes not grammatically correct, making them hard to embed in BERT for automatic scoring. 

As an example of LLMs, ChatGPT naturally has the benefits and drawbacks outlined above. However, many of these difficulties can be resolved as ChatGPT is open to be fine-tuned. Prior research (e.g., \citep{wu2023matching}) has presented how fine-tuned models can be made to do particular tasks, such as computerized grading in science education. More precise, equitable, and domain-appropriate automatic scoring can be achieved by aligning ChatGPT with corpora specific to a given domain and fine-tuning based on context. Furthermore, the model's dependability and credibility in educational environments can be further increased by iterative fine-tuning based on feedback loops with educators \citep{liu2023context}. Fine-tuned GPT models are more suited to tasks like text completion, response evaluation, or open-ended queries because of their autoregressive nature, which excels in sequence formation. GPT's already strong generating capacities will be strengthened by fine-tuning it for specific tasks or domains. This will enable it to produce more contextually relevant and accurate outputs, effectively solving some of the limitations of BERT.

\textcolor{black}{A recent advancement to Bard \citep{bard2023google}, Gemini Pro, a multimodal, which shows the technical competition for state-of-the-art (SOTA) AI services. Google Bard and Gemini Pro have also been tested and used for reasoning, answering knowledge-based questions, solving math problems, translating between languages, generating code, and acting as instruction-following agents through SOTA benchmarks \citep{gemini2023google}. However, Google only released its APIs\footnote{\url{https://ai.google.dev/tutorials/python\_quickstart}} on 13th December 2023 to create applications for Gemini Models; it still doesn't provide public APIs to fine-tune the models. Hence, our study has to restrict our findings to GPT3.5 as we focus on providing capabilities of fine-tuned GPT-3.5 Turbo for automatic scoring; hence, the publicly released Bard is not the right fit for the study.}


\section{Methodology}\label{methodology}
\subsection{Datasets}
This study is a secondary analysis of existing datasets, consisting of responses from middle school students with expert scoring for two multi-label assessment tasks from the PASTA project \citep{ Harris2024Creating, PASTA2023} and four multi-class assessment tasks from the Mathematical Thinking in Science (MTS) project \citep{Jin2019A, ETS2023} responded to by high-school students. The data division based on each assessment item can be seen in Table.~\ref{table:data}.
\begin{table}[hbt!]
    \centering
    \caption{Dataset information for both multi-label and multi-class tasks}
    \begin{tabular}{p{2cm} l l l l l}
        \toprule
        \textbf{Model Type} &\textbf{Item}& \textbf{No.} & \textbf{Labels/Classes} & \textbf{Training size} & \textbf{Testing size} \\
        \midrule
        Multi-label & Gas-filled balloons & 5 & 5 binary PEs & 1200 & 240 \\
        & Layers in test tube & 10 & 10 binary PEs & 1428 & 285 \\
        
       Multi-class & Falling weights & 4 & (0-3) level& 1150 & 230 \\
        & Gelatin & 5 & (0-4) level & 1148 & 226 \\
        & Bathtub & 5 & (0-4) level & 1145 & 230 \\
        & Sandwater1 & 4 & (0-3) level & 1144 & 230 \\
        \bottomrule
    \end{tabular}
    \label{table:data}
\end{table}

\subsection{Assessment Task}
\subsubsection{Multi-label Assessment Tasks}
The assessment tasks are designed to assess the middle school student's ability to use multi-label knowledge to explain scientific phenomena. Aiming to support students in developing knowledge-in-use across grade levels, NGSS provides performance expectations for K–12 students that integrate disciplinary core ideas (DCIs), crosscutting concepts (CCCs), and science and engineering practices (SEPs). The tasks employed in this study are aligned with the NGSS performance expectation at the middle school level: Students analyze and interpret data to determine whether substances are the same based on characteristic properties \citep{national2013next}. This performance expectation requires students to be able to use the structure and properties of matter and chemical reactions (DCIs) and patterns (CCC) to analyze and interpret data (SEP).
\begin{figure}[hbt!]
    \centering
    \includegraphics[width=1\linewidth]{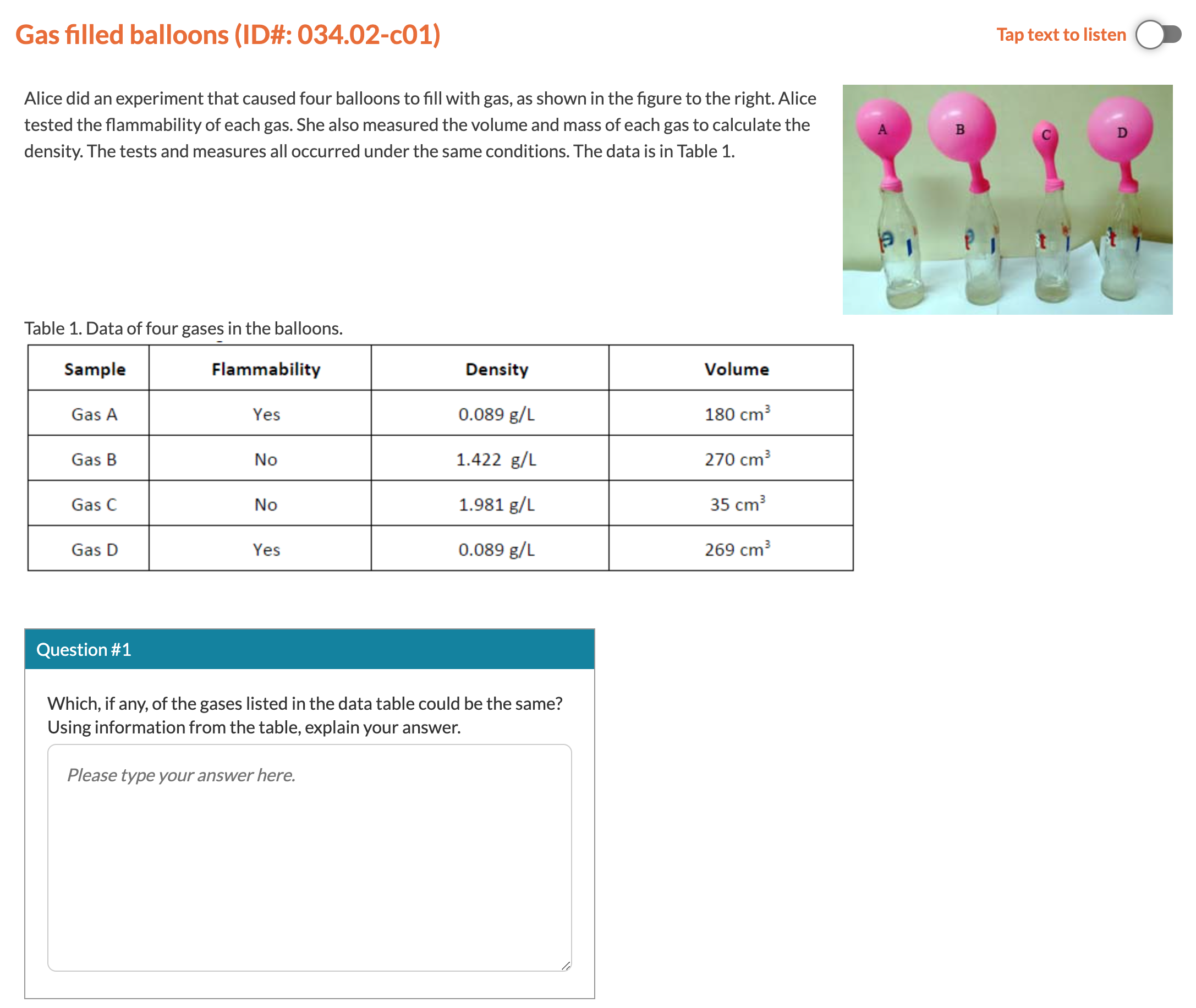}
    \caption{Example Multi-label Task: Gas-Filled Balloons}
    \label{fig:gas_filled_ballons}
\end{figure}
1200 students from grades 6-8 participated in this study. Middle school teachers in the US were offered the opportunity for their students to test open-ended NGSS-aligned science tasks \citep{zhai2022applying}. Student responses from this system were randomly selected as a representative participant set for the machine learning algorithms' training, validation, and testing. Student data was masked in this sample to protect student privacy; thus, no demographic information is available to researchers. However, given the geographic distribution of the teachers choosing to access the system, the data set is considered a representative cross-section of US middle school students.

These assessment tasks were selected from the Next Generation Science Assessment \citep{Harris2024Creating}. These questions dealt with applying \textit{chemistry} ideas in real-world situations. These tasks come under the physical sciences category of "Matter and its Characteristics," which assesses students' analytical and interpreting skills to compare different substances based on their distinctive properties. The tasks aim to examine students' multi-perspective learning and give teachers the information they may utilize to help decision-making in the classroom. The automatic reports from the students' rubric scores show what subjects they still need to learn more. 

For example, In the first task, students were asked to identify gases in the experiment based on how closely their attributes match those listed in the data table (see Fig.~\ref{fig:gas_filled_ballons}). To answer this question, students have to understand the structure and properties of matter and chemical reactions and be able to use the knowledge to plan and carry out investigations and identify patterns in the data. 

We developed a scoring rubric containing five response perspectives based on the dimensions of science learning: SEP+DCI, SEP+CCC, SEP+CCC, DCI, and DCI. This scoring rubric reflects multi-perspective thinking \citep{He2024G}. Table~\ref{table:rubric_gas_filled_ballon} lists specific components for each dimension. Students would receive multiple scores automatically and simultaneously based on a multi-perspective (MPS) rating if they understood the DCIs, CCCs, or SEPs aligned with the rubric. The research team collaborated with science teachers to further validate the multi-perspective rubrics. 

\begin{table}[hbt!]
\centering
\caption{Scoring rubric for task: Gas-filled balloons.}
\begin{tabular}{l l p{9.5cm} }
\toprule
\textbf{ID} & \textbf{Perspective} & \textbf{Description} \\
\midrule
E1 & SEP+DCI &  Student states that Gas A and D could be the same substance. \\
E2 & SEP+CCC & Student describes the pattern (comparing data in different columns) in the table flammability data of Gas A and Gas D as the same. \\
E3 & SEP+CCC & Student describes the pattern (comparing data in different columns) in density data of Gas A and Gas D, which is the same in the table. \\
E4 & DCI & Student indicate flammability is one characteristic of identifying substances. \\
E5 & DCI & Student indicate density is one characteristic of identifying substances. \\
\bottomrule
\end{tabular}
\label{table:rubric_gas_filled_ballon}
\end{table}

\subsubsection{Multi-Class Assessment Tasks}
The multi-class assessment tasks ask students to solve a science problem using scientific knowledge and mathematical reasoning (e.g., proportional reasoning) \citep{jin2019validation}.  The tasks intend to assess the nature of mathematical reasoning in science and how students acquire this ability as they progress through high school science courses. The tasks target two NGSS disciplinary core ideas: PS3 (Energy) and LS2 (Ecosystems: Interactions, Energy, and Dynamics) at the high school level.  Approximately 1400 students from grades 9-12 participated in this study, and eight teaching experts graded the responses based on the respective rubric.

 To respond to these tasks, students should consider the following concepts:
\begin{enumerate}
    \item Heat ($Q$) and temperature ($T$): Heat is an extensive variable that depends on the amount of substance (mass, M), while temperature is an intensive variable that does not depend on the amount of substance. Lower-level performing students often are confused with heat and temperature. They often think that temperature is the measure of heat (and not the average energy of molecular motion in a substance). They may also use a single undifferentiated variable that conflates the meaning of heat and the meaning of temperature to explain phenomena. In contrast, the higher-performing students understand that heat and temperature are two different quantities.
    \item Specific Heat as a compound concept: Specific heat is the amount of heat required to raise the temperature of a unit of mass of a substance by one degree Celsius. The scientific meaning of specific heat emphasizes that specific heat is a substance’s internal property to regulate temperature. The material determines this internal property is the type of material or the material's microscopic/atomic/molecular structure. Specific heat is a compound concept because it results from calculating other variables.  Lower-performing students may only recognize mathematical patterns in diagrams of $Q-T$ but do not understand how specific heat is related to the slope $\left(C = \frac{Q}{M\Delta T}\right)$.  
    \item For the same amount of $Q$, $\Delta T$ depends on both $M$ and $C$: Heat input/output will cause the temperature of a substance to increase/decrease. For a certain amount of heat, the temperature changes depending on two factors—the amount of the substance (i.e., mass, $M$) and the type of the substance (i.e., heat capacity, $C$).  
\end{enumerate}

\begin{figure}[hbt!]
    \centering
    \includegraphics[width=1\linewidth]{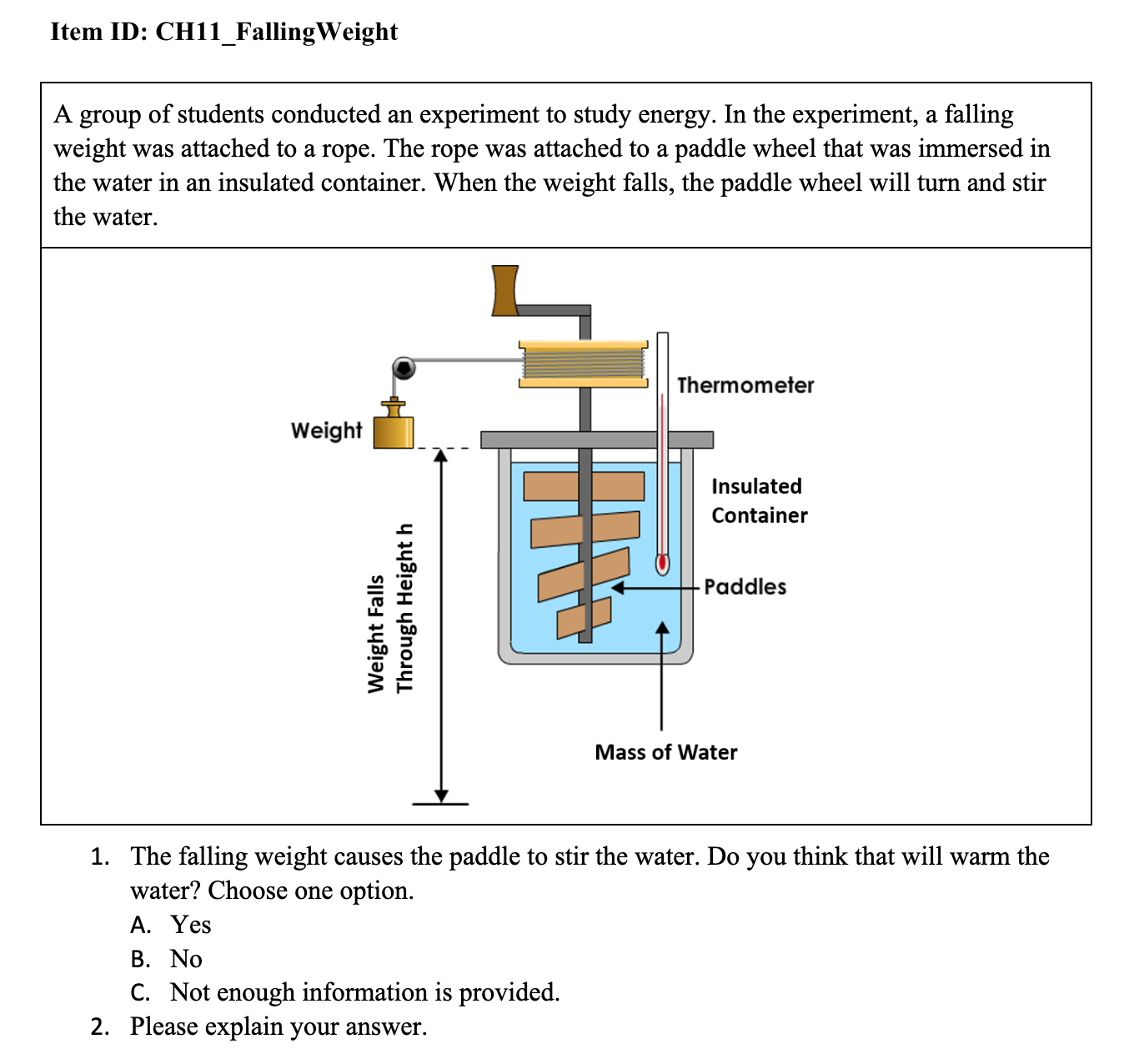}
    \caption{Example Multi-class Task: Falling weights}
    \label{fig:falling_weights}
\end{figure}

One example of a multi-class task shows a phenomenon in which a falling weight drives a paddle wheel to stir the water. Students are asked to determine whether the water will turn warm and write an explanation. For this example task, students have to identify the heat change in water based on falling weight, which causes the paddle to stir the water(see Fig.~\ref{fig:falling_weights}).

To assess students' knowledge use proficiency in responding to the example item, a four-level rubric was developed. Table~\ref{table:rubric_falling_weights} lists specific descriptions for each level. 


\begin{table} [hbt!]
\centering
\caption{Scoring rubric for task: Falling weights}
\begin{tabular}{l p{5.5cm} p{5.5cm} }
\toprule
\textbf{Level} & \textbf{Description} & \textbf{Example} \\
\midrule
3 & A Level 3 response suggests that the student understands the measurability of variables. Level 3 responses must choose A. There are two patterns at Level 3: & Example 1: A [Yes.] Because the weight will stir the water, the movement in water particles will cause the temperature to rise. \\
& \textbf{3a:} [Identification of heat/energy] The response 1) associates movement (e.g., movement of the weight, paddle, and/or water) with energy/heat AND/OR 2) associates particle movement with temperature, heat, or energy. & Example 2: A. [Yes.] The falling weight transfers its energy to the paddles that spin. The spinning paddles transfer their energy to the water. Because water (the system) absorbs energy, the temperature of the water will increase even if it is very small. \\
& \textbf{3b:} [Identification of heat/energy with errors.] The response 1) associates movement (weight, paddle, and/or water) with energy or heat AND/OR 2) associates particle movement with temperature, heat, or energy. But, the response also confuses energy/heat/temperature with other variables such as forces. & Example 3: A. [Yes.] The weight falling will cause the paddle to start moving, which turns gravity into kinetic energy, and the kinetic energy is then transferred to the water and moves the water's molecules, which would then start heating the water. \\

2 & A Level 2 response: & Example 1: C [Not enough information is provided.] It depends on how fast the paddles are turning. If they are turning fast enough, they could warm the water.\\
& \textbf{2a}: [Threshold] The response indicates a threshold where the movement must be fast to a certain degree or the energy input must be large enough to cause an increase in temperature. & Example 2: B. [No.] Stirring the water will not increase the temperature of the water because no energy is being added. \\
& \textbf{2b:} [General understanding] The response indicates that the student generally understands that work/energy will cause the temperature to increase but cannot apply that general understanding to this specific scenario. & Example 3. C. [Not enough information is provided.] there is not enough information to explain more about the temperature; just will it spin, and yes, it will spin pretty fast. \\
& \textbf{2c:} [Macroscopic causation] The response provides a causal relationship at a macroscopic scale without using energy or heat. & Example 4: B. [No.] We don't know if the water is warm or cold already, room temperature, etc.\\
& \textbf{2d:} [Irrelevant variables ONLY] The response analyzes the scenario based on irrelevant variables. & Example 5: A. [Yes] Due to the friction between the paddle and the water, the water will get warmer after some time.\\
1 & Level 1 responses have the following patterns: & Example 1: A. [Yes.] It's placed in an insulated container, and it has a thermometer. \\
& \textbf{1a:} [IDK] "I don’t know" type of response. &  \\
& \textbf{1b:} [No information] The response does not provide information about the student’s ideas about the phenomenon or data. &  \\
0 & Blank or random letters & Acareba artouth \\
\bottomrule
\end{tabular}
\label{table:rubric_falling_weights}
\end{table}

\subsection{Machine Algorithms}
\subsubsection{BERT}
The cutting-edge language representation model BERT (Bidirectional Encoder Representations from Transformers) was developed by \citep{devlin2018bert} based on the Transformer model \citep{vaswani2017attention}. One of its key advances is the Transformer model's self-attention mechanism, which compares the significance of several words in a phrase to a particular word. In contrast to conventional models, which usually concentrate in one direction, this method enables the model to collect context in both the forward and backward directions. Due to its bi-directionality, BERT can better comprehend the precise meaning of each word in a phrase.

BERT's in-depth comprehension of context can be quite useful for automatic scoring. For instance, BERT's capacity to comprehend complex linkages between many response components becomes essential in evaluating student failures in experimentation. The effectiveness of Large Language Models in evaluating student experimentation errors was compared with that of human raters in a study by \citep{bewersdorff2023assessing}. Their findings suggested that models like BERT, which have a deep awareness of context, could nearly mimic the evaluation abilities of human experts in some science education. Similar research that employed BERT for automatic scoring of student written responses in science \citep{riordan2020empirical} reported robust accuracy.

Given the outstanding performance of BERT, compared to other machine learning models, in the automatic scoring of written responses in science education, we used BERT as a baseline.

\subsubsection{GPT-3.5}
Modern language model GPT-3.5, created by OpenAI \citep{brown2020language} as a variation of the Generative Pre-trained Transformer (GPT) series, is a state-of-the-art LLM. The Transformer that  uses the self-attention mechanism \citep{ghojogh2020attention} to balance the importance of various words in a sequence forms the foundation of GPT-3.5's internal architecture. This model offers a potent representation of contextual relationships.

The potential applications of GPT-3.5 go beyond straightforward text production. It can adapt to various tasks without requiring tedious fine-tuning because of its few-shot learning capabilities. Translation, summarizing, answering questions, and even programming duties fall under this category \citep{floridi2020gpt}. The analysis by Floridi clarifies the nature of GPT-3.5, highlighting its ability to function as an agency even though it lacks accurate intelligence and emphasizes the creative power of such massive models. \citep{kung2023performance} demonstrated the model's success on the USMLE (United States Medical Licensing Examination) to illustrate its potential in AI-assisted medical education. Researchers can even use GPT to assist in writing academic papers in education \citep{zhai2022chatgpt}. GPT-3.5 is a strong choice for automatic scoring in educational settings because of its few-shot learning capacity. Traditional automatic scoring systems need a large number of labeled domain-specific datasets to fine-tune the algorithm. The requirement for domain-specific fine-tuning, however, is lessened by GPT-3.5. \citep{zhai2023chatgpt} found that GPT-3.5 shows great potential for automatically scoring written responses to science equations. The model can align its scoring pattern with human raters by comprehending the scoring requirements and evaluating responses with minimum input \citep{wang2022self}.

Additionally, because of its ability to comprehend and produce human-like language, it can provide feedback on student responses that include both a score and qualitative observations, or even customized learning guidance based on students' performance \citep{zhai2023chatgpt}. GPT-3.5 and comparable models would be feasible to acquire with the democratization of huge language models espoused by \citep{candel2023h2ogpt}, making AI-driven automatic scoring a common occurrence in educational institutions.

GPT-3.5's effectiveness in situations requiring automatic scoring can significantly improve with minor adjustments. GPT-3.5 can be modified to capture the nuances and intricacies of the scoring criteria by altering the model's weights on a particular scoring dataset, leading to enhanced alignment with human raters' expectations. The InstructionGPT-4 study by \citep{wei2023instructiongpt} shows how big models may be tuned using a paradigm of 200 instructions, resulting in more accurate task-specific performance. Such methods could significantly reduce scoring differences between humans and artificial intelligence (AI), guaranteeing that student responses are evaluated fairly and consistently for the GPT-3.5.

Overall, GPT-3.5 is a robust candidate for the development of autonomous scoring due to its architecture and capabilities. It has great potential to transform the field of automatic assessment thanks to its in-depth context awareness and flexibility. We have used a refined version (GPT-3.5-turbo) from the family of GPT-3.5 for fine-tuning because of its API availability. GPT-3.5-Turbo is an improved version of GPT-3 and GPT-3.5 to balance performance and efficiency. GPT-3.5-Turbo has capabilities comparable to GPT-3 but with fewer settings. This makes its use in a variety of applications affordable.

\subsubsection{Domain-Specific Training}
Large language models like GPT-3.5-turbo that have undergone domain-specific training can produce even more precise outcomes adapted to certain situations, topics, or scoring criteria. Text data augmentation is one creative way to use GPT-3.5-turbo's capabilities for this objective. A larger and more varied training set for scoring mechanisms can be produced using GPT-3.5 to supplement textual datasets, as demonstrated by \citep{cochran2023improving}. By enhancing the model's comprehension and offering a wider variety of sample responses, such augmentation has been proven to improve the performance of automated evaluation of student text responses.

The PandaLM benchmark was introduced by \citep{wang2023pandalm} to optimize domain-specific training. This evaluation environment is specifically for adjusting and optimizing LLM instructions. The benchmark enables a thorough understanding of how well the model performs across various tasks, identifying potential improvement areas and fine-tuning strategies to increase the model's effectiveness in domain-specific applications, such as automatic scoring.

To sum up, domain-specific training of GPT-3.5-turbo paves the way for more accurate and consistent automated scoring systems and ensures that the model grows more adept at comprehending the intricate aspects of particular subjects or tasks.

\section{Experimental Setup}\label{experiments}
\subsection{Training Scheme}
For their outstanding generalization across a broad range of applications, LLMs like GPT-3.5-turbo have shown great potential. However, optimizing these models on domain-specific datasets for specialized domains like education is crucial, particularly in the context of automated scoring of student responses. Here, we outline our training scheme for optimizing GPT-3.5-turbo using academic data.

\subsubsection{Data Collection and Preprocessing}
A thorough dataset with student replies is necessary to start with. To guarantee a broad training set, these responses should include a range of themes, question kinds, and levels of complexity. Each response should have a human-assigned grade or score to give the model a clear supervisory signal. \textcolor{black}{Following steps are applied for data collection and processing while considering OpenAI data processing guidelines\footnote{\url{https://platform.openai.com/docs/guides/fine-tuning/preparing-your-dataset}}:}
\begin{itemize}
    \item \textbf{Data Cleaning:} Remove any irrelevant or personally identifiable information from the dataset to maintain student privacy.
    \item \textbf{Tokenization:} Convert the textual responses into a format suitable for the model using appropriate tokenization strategies.
    \item \textcolor{black}{\textbf{Uploading Files:} After processing the dataset, JSON files are generated and can be uploaded to the OpenAI server using file upload API\footnote{\url{https://platform.openai.com/docs/api-reference/files}}}.
\end{itemize}

\subsubsection{Model Initialization}
The pre-trained GPT-3.5-turbo model needs to be loaded first. Starting with a previously trained model, such as GPT-3.5-turbo, has the advantage of understanding the language's structure, which speeds up convergence during fine-tuning.

\subsubsection{Fine-tuning Procedure}
\begin{enumerate}
    \item \textbf{Loss Function:} Use a regression-based loss function if the scores are continuous or a classification loss if the scores are categorical.
    \item \textbf{Learning Rate:} Start with a small learning rate to ensure that the fine-tuning process doesn't diverge from the pre-trained weights too aggressively.
    \item \textbf{Epochs:} Depending on the dataset's size, several epochs might be required to achieve convergence. Monitor the validation loss to prevent overfitting.
    \item \textbf{Batch Size:} Choose an appropriate batch size, considering the available computational resources.
\end{enumerate}

\subsubsection{Evaluation and Validation}
Once the model is fine-tuned, an evaluation set (distinct from the training set) is necessary to assess the model's performance.
\begin{itemize}
    \item \textbf{Metrics:} Depending on the scoring nature, metrics like Mean Absolute Error (MAE) or classification accuracy can be used to gauge the model's scoring precision against human raters.
    \item \textbf{Comparison with Baseline:} Compare the fine-tuned model's performance with the original GPT-3.5-turbo to quantify the benefits of domain-specific fine-tuning.
\end{itemize}



\subsection{Baselines}
This study intends to investigate how the training data context affects the performance of pre-trained models (like BERT) and to investigate methods to enhance model performance further. To accomplish this goal, we use a variety of datasets to train and improve the models. First, we use 7T as the downstream task and the original BERT as the pre-trained model, which serves as the standard model. Then, using 7T as the downstream task, we train a BERT model using the dataset from each task. We contrast the two refined models to meet our goals, as the 7T is based on the context of science education.

\section{Results}

\begin{figure}
    \centering
    \includegraphics[width=1\linewidth]{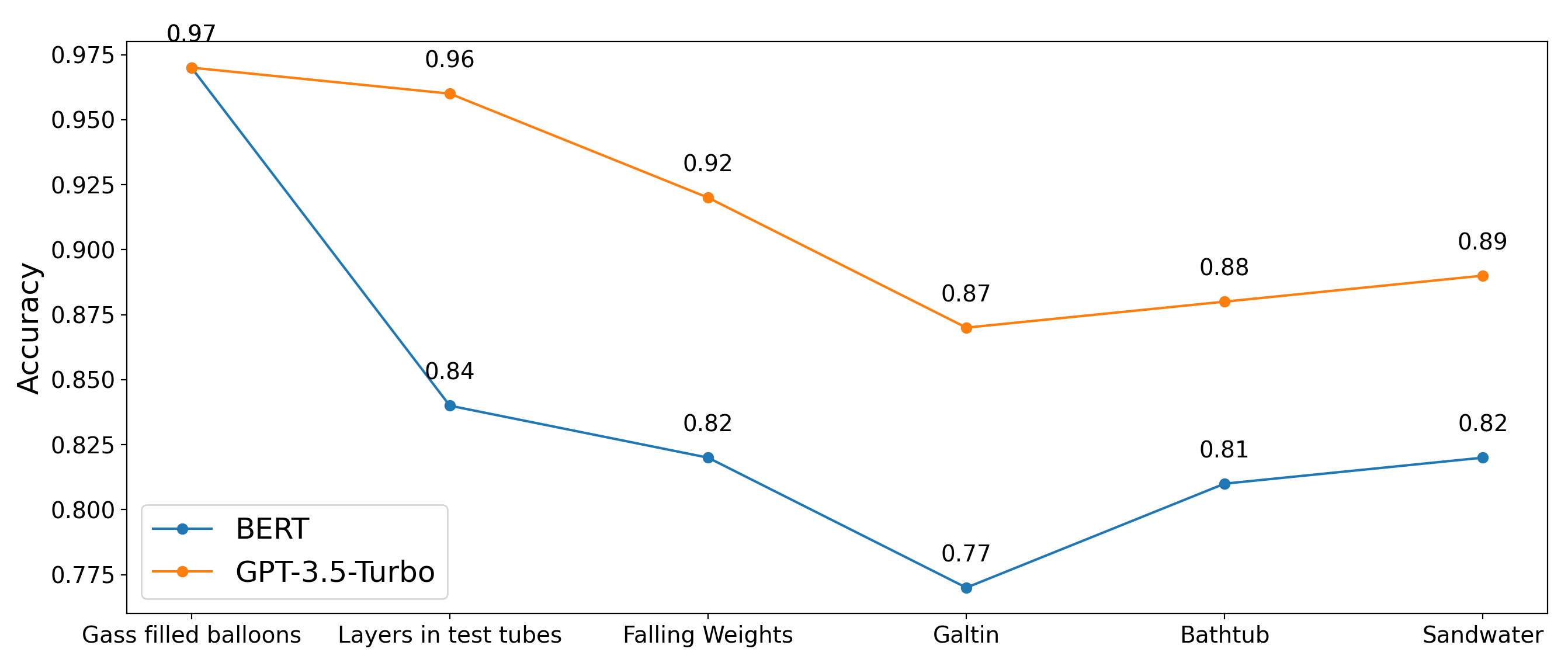}
    \caption{Accuracy comparison of fine-tuned GPT-4.5-Turbo and BERT for different assessment tasks}
    \label{fig:accuracy_plot}
\end{figure}

We performed a paired-sample T-test to compare the accuracies of fine-tuned GPT-3.5-turbo and BERT. We also calculated the p-value for the two-tailed test as we are interested in analyzing the accuracy difference from both directions. Across the trained and validated datasets yielded significant gains for GPT-3.5-turbo over BERT (Mean Difference = 0.076, SD = 0.043) conditions; \textit{t}(5) = 4.44, \textit{p} $=0.007 < 0.05$ \citep{kahn2010reporting}. The results indicate that the average accuracy of fine-tuned GPT-3.5-turbo is significantly higher than that of the fine-tuned BERT, with 9.1\% increase.
A direct accuracy comparison plot can be seen in Fig.~\ref{fig:accuracy_plot}.

\subsection{Multi-label Tasks}
Two categories were evaluated for the multi-label tasks dataset: 'Gas-filled balloons' and 'Layers in test tube'.

\begin{itemize}
    \item \textbf{Gas-Filled Balloons:} Both BERT and GPT-3.5-turbo achieved a remarkable accuracy of $0.97$ in this multi-label classification task. No significant difference was observed between BERT and GPT-3.5-turbo, \textit{t}(2) = 0.  This suggests that the two models are equally adept at identifying and categorizing elements related to gas-filled balloons.
    \item \textbf{Layers in Test Tubes:} A significant difference of $0.12$ was noted, with GPT-3.5-turbo outperforming BERT. The significant performance boost indicates GPT-3.5-turbo's heightened capability in grasping the complexities associated with this category.
\end{itemize}
As both tasks are multi-labeled, we also performed a paired-sample t-test to compare the individual label's accuracy for fine-tuned GPT-3.5-turbo and BERT.  A significant difference was observed for both tasks. For the first task, a Mean Difference = 0.008 (SD = 0.008) of scoring accuracy between GPT-3.5-turb and BERT   was observed,  \textit{t}(4) = 2.14 and \textit{p} $=0.001 < 0.05$. For the second task, which has more unbalanced labels (Mean Difference = 0.065, SD = 0.044)  conditions; \textit{t}(9) = 4.67 and \textit{p} $=0.001 < 0.05$. Upon examining the multi-label dataset, particularly the 'Layers in Test Tubes' category, it's evident that GPT-3.5-turbo consistently achieved higher accuracy on certain unbalanced labels than BERT. This consistent performance across varying label distributions underscores GPT-3.5-turbo's capability to effectively handle unbalanced labels in multi-label datasets, offering a clear advantage (7.2\% increase) over models like BERT in specific scenarios.

\subsection{Multi-class Tasks}
We also performed a paired-sample t-test to compare the accuracy of fine-tuned GPT-3.5-tubo and BERT for all the multi-class tasks. There was a significant difference in the accuracies between the two models (Mean Difference = 0.085, SD = 0.017)  conditions; \textit{t}(4) = 9.82 and \textit{p} $=0.0006 < 0.05$. The results suggest that GPT-3.5-turbo can predict scores for multi-class tasks significantly more accurately, with an average increase of 10.6\%. Specifically,  
\begin{itemize}
    \item \textbf{Falling Weights:} GPT-3.5-turbo outperformed BERT with an accuracy of $0.92$ compared to BERT's $0.82$. This indicates GPT-3.5-turbo's superior ability to distinguish between different classes related to falling weights.
    \item \textbf{Gelatin:} Once again, GPT-3.5-turbo took the lead with an accuracy of $0.87$, a notable improvement from BERT's $0.77$. The results highlight GPT-3.5-turbo's refined understanding of concepts tied to gelatin.
    \item \textbf{Bathtub:} Both models performed commendably in this category, with GPT-3.5-turbo scoring $0.88$ and BERT trailing slightly at $0.81$. The performance differential suggests the advanced capabilities of GPT-3.5-turbo in discerning nuances related to bathtubs.
    \item \textbf{Sandwater1:} Here too, GPT-3.5-turbo surpassed BERT, scoring $0.89$ against BERT's $0.82$. This underscores GPT-3.5-turbo's adeptness at handling concepts associated with 'sandwater1'.
\end{itemize}

It is evident from the findings that GPT-3.5-turbo consistently outperformed BERT, particularly in the multi-class tasks. The architecture of GPT-3.5-turbo, the large amount of training data, or its innate skills to comprehend context more effectively may be responsible for its continual performance improvement. While GPT-3.5-turbo has shown higher proficiency across various educational categories, making it a preferable option for automatic scoring in certain scenarios, BERT has displayed commendable performances, particularly with "Gas-filled balloons."

\section{Discussion}\label{discussion}
Automatic scoring has been long regarded as essential for effective assessment practices in education. Researchers have applied various machine learning algorithms and strategies to achieve high-scoring accuracy \citep{zhai2020applying}, though with varying degrees of success \citep{zhai2021meta}. The results from comparing accuracies of fine-tuned BERT and GPT-3.5-turbo in automatic scoring provide informative information about the potential of using fine-tuned GPT for automatic scoring in educational contexts. Pre-trained models can be effectively adapted to tasks specifically to a certain domain using the optimization technique known as fine-tuning. According to our results in this study, instruction-based fine-tuning can improve accuracy in tasks like automatic scoring for both multi-label and multi-class scoring tasks. Thanks to this domain-specific adaption, the fine-tuned scoring models can capture the complexities and intricacies unique to the datasets in education.

Our results provide evidence of GPT-3.5-turbo's improved performance in automatic scoring of student written responses, highlighting the technology's potential as a cutting-edge tool for educational applications. Prior research has demonstrated the potential of large-language models in automatic scoring of student written responses \citep{liu2023context,riordan2020empirical}, the architecture of GPT-3.5-turbo further shows its capacity to comprehend context more holistically compared to existing LLMs, and its training strategy tailored for educational data contribute to the consistent and remarkable performance gains. The performance is consistent with expectations for ChatGPT due to its powerful language analysis and generation capacity.

The availability of fine-tuned ChatGPT is a huge advance for specific tasks such as automatic scoring in education. Our findings show the promise of GPT-3.5-turbo in education, where evaluating student responses calls for both accuracy and context knowledge. Its fine-tuned version, created especially for educational data, allows scoring procedures to be automated without sacrificing accuracy. The findings show that the GPT-3.5-turbo can be a beneficial tool in educational technology, opening the door for faster and more reliable evaluations with the proper training and fine-tuning techniques.

While the fine-tuned GPT-3.5-turbo achieved outstanding results compared to BERT, it also shows the promise of minority scoring categories in unbalanced data. Data unbalance has been a significant concern in machine training and finetuning processes \citep{zhai2023ai}, it is thus expected that the more advanced algorithms can contribute to this gap. Our findings show that in "Gas-filled balloons," the GPT-3.5-turbo has a clear advantage in minority scoring categories, demonstrating its prowess in understanding various educational content. 

Overall, the rise of models like GPT-3.5-turbo and their potential after fine-tuning signifies a transformative shift in the automatic scoring domain. The promising results achieved by GPT-3.5-turbo, especially compared to BERT's performance, reinforce the importance of leveraging advanced models for educational applications. As models evolve, the potential for more refined and domain-specific applications in education is vast, warranting further exploration and research. 

\section{Conclusions and Limitations}
\label{conclusion}
This study highlights the potential of ChatGPT, notably GPT-3.5, for automatic scoring in science education. In our study, we focused on multi-label and multi-class assessment tasks to fine-tune GPT-3.5, utilizing varied datasets that included responses from middle-school and high-school students. The study closes the knowledge gap between the pre-trained model and the contextual demands of the educational sector. Our finding suggests that for automatic scoring accuracy, fine-tuned GPT-3.5 outperformed the refined version of Google's cutting-edge language model, BERT. This finding demonstrates GPT-3.5's effectiveness in automating scoring tasks for students' responses specific to a given domain and presents a scalable and replicable approach to utilizing this capability for various science education tasks that require pronounced accuracy. This study represents a proof-point for the expanding educational sector potential of fine-tuned language models and a lighthouse for future initiatives trying to tap into this potential for various educational applications.

Despite of the significant findings, the study has limitations. First, as a secondary analysis of existing assessment responses, researchers were constrained to the assessment process and specific student and teacher information. Future research should may focus on the implementation of automatic scoring in authentic classroom settings and examine the student learning outcomes using experimental design. 

Second, we acknowledge that a boom of LLMs emerged in the market, such as the recently released Google's Gemini Pro \citep{gemini2023google} incorporated in Bard \citep{bard2023google}, and our study has only compared two SOTA LLMs (BERT and GPT-3.5) for fine-tuning and automatic scoring. Once the open-source and public API-provided LLMs for fine-tuning and Google APIs\footnote{\url{https://ai.google.dev/}}  is released, we suggest that future research should compare GPT-3.5 turbo with Gemini and other similar models for automatic scoring in education.

Third, integrating AI into educational assessment, specifically fine-tuned ChatGPT models, raises ethical dilemmas. 
Despite the advancement of automatic scoring \citep{latif2023automatic} in providing detailed feedback, poses questions about the fairness and transparency of the scoring algorithms need to be addressed \citep{latif2023ai}. Future research should further study these ethical issues to better implement automatic scoring in classroom settings.

Fourth, the availability of automatic scoring has drawn concerns about teachers' roles in the era of AI-assisted education. The use of ChatGPT for providing feedback and scoring might alter the traditional role of educators, shifting their focus from assessment to more personalized guidance and support \citep{adiguzel2023revolutionizing}. Furthermore, the efficiency and ease of training and uploading tests for AI-based assessment compared to traditional human marking are undeniable. However, this efficiency comes with concerns regarding the potential reduction in critical human engagement in the learning process. The reliance on AI for feedback and scoring might lead to a diminished capacity for critical thinking and personal interaction, which are crucial components of effective education \citep{mhlanga2023open}. Future research should examine teachers' pedagogical needs to navigate the teacher-AI relationships in educational processes to better serve student learning \citep{halaweh2023chatgpt}.

Last, despite the promise of automatic scoring, the use of ChatGPT in educational settings must be scrutinized for issues related to bias, privacy, and data security \citep{huallpa2023exploring}. Ensuring that the AI systems are free from inherent biases and that student data is handled responsibly is paramount to maintaining ethical standards in education.



\section*{Acknowledgment}
This study secondary analyzed data from projects supported by the National Science Foundation (grants numbers 
\#) and the Institute of Education Sciences (grant number 
). The authors acknowledge the funding agencies and the project teams for making the data available for analysis. The findings, conclusions, or opinions herein represent the views of the authors and do not necessarily represent the views of personnel affiliated with the funding agencies. 



\section*{Declaration of Human Subject Approval}
According to U.S. federal regulations, human subject research involves obtaining data through interaction with individuals or identifiable private information about them. If the data is completely de-identified and the researcher cannot link it back to individual subjects, it may not be considered human subject research. In our case, the researchers secondarily analyzed existing de-identified data, which have undertaken IRB review. This specific research is not regarded as human subject research and thus is waived from further IRB review.
\color{black}

\section*{Declaration of generative AI and AI-assisted technologies in the writing process}

During the preparation of this work the author(s) used ChatGPT in order to check grammar and polish the wordings. After using this tool/service, the authors reviewed and edited the content as needed and take full responsibility for the content of the publication.

\bibliography{sn-bibliography}

\end{document}